\documentclass[pdflatex,sn-mathphys-num]{sn-jnl}% Math and Physical Sciences Numbered Reference Style 
\usepackage{graphicx}%
\usepackage{multirow}%
\usepackage{amsmath,amssymb,amsfonts}%
\usepackage{amsthm}%
\usepackage{mathrsfs}%
\usepackage[title]{appendix}%
\usepackage{xcolor}%
\usepackage{textcomp}%
\usepackage{manyfoot}%
\usepackage{booktabs}%
\usepackage{algorithm}%
\usepackage{algorithmicx}%
\usepackage{algpseudocode}%
\usepackage{listings}%
\theoremstyle{thmstyleone}%
\theoremstyle{thmstyletwo}%

\theoremstyle{thmstylethree}%

\begin{document}

\title[Article Title]{Is AI Robust Enough for Scientific Research?}


\author[1]{\fnm{Jun-Jie} \sur{Zhang}}\email{zjacob@mail.ustc.edu.cn}

\author[11]{\fnm{Jiahao} \sur{Song}}\email{jiahaosong@mail.nwpu.edu.cn}
\equalcont{These authors contributed equally to this work.}

\author[2,3]{\fnm{Xiu-Cheng} \sur{Wang}}\email{xcwang\_1@stu.xidian.edu.cn}
\equalcont{These authors contributed equally to this work.}

\author[4]{\fnm{Fu-Peng} \sur{Li}}\email{fupengli29@mails.ccnu.edu.cn}
\equalcont{These authors contributed equally to this work.}

\author[10]{\fnm{Zehan} \sur{Liu}}\email{2311531808@qq.com}
\equalcont{These authors contributed equally to this work.}

\author[1]{\fnm{Jian-Nan} \sur{Chen}}\email{chennn1994@alumni.sjtu.edu.cn}

\author[15]{\fnm{Haoning} \sur{Dang}}\email{haoningdang.xjtu@stu.xjtu.edu.cn}

\author[9]{\fnm{Shiyao} \sur{Wang}}\email{wang.s.be@m.titech.ac.jp}

\author[5]{\fnm{Yiyan} \sur{Zhang}}\email{zhangyy414@stu.xjtu.edu.cn}

\author[8]{\fnm{Jianhui} \sur{Xu}}\email{xujianhui306@gdas.ac.cn}

\author[7]{\fnm{Chunxiang} \sur{Shi}}\email{shicx@cma.gov.cn}

\author[15]{\fnm{Fei} \sur{Wang}}\email{feiwang.xjtu@xjtu.edu.cn}

\author[4]{\fnm{Long-Gang} \sur{Pang}}\email{lgpang@ccnu.edu.cn}

\author[2,3]{\fnm{Nan} \sur{Cheng}}\email{dr.nan.cheng@ieee.org}

\author[11]{\fnm{Weiwei} \sur{Zhang}}\email{aeroelastic@nwpu.edu.cn}

\author[12,13,14]{\fnm{Duo} \sur{Zhang}}\email{zhduodyx@pku.edu.cn}

\author*[6,15]{\fnm{Deyu} \sur{Meng}}\email{dymeng@mail.xjtu.edu.cn}

\affil[1]{\orgname{Northwest Institute of Nuclear Technology}, \orgaddress{\street{No. 28 Pingyu Road}, \city{Xi'an}, \postcode{710024}, \state{Shaanxi}, \country{China}}}

\affil[2]{\orgdiv{School of Telecommunications Engineering}, \orgname{Xidian University}, \orgaddress{\street{No. 2 South Taibai Road}, \city{Xi'an}, \postcode{710071}, \state{Shaanxi}, \country{China}}}
\affil[3]{\orgname{State Key Laboratory of ISN}, \orgaddress{\street{No. 2 South Taibai Road}, \city{Xi'an}, \postcode{710071}, \state{Shaanxi}, \country{China}}}

\affil[4]{\orgdiv{Key Laboratory of Quark and Lepton Physics (MOE) \& Institute of Particle Physics}, \orgname{Central China Normal University}, \orgaddress{\street{No. 152 Luoyu Road}, \city{Wuhan}, \postcode{30079}, \state{Hubei}, \country{China}}}

\affil[5]{\orgdiv{School of Computer Science and Technology}, \orgname{Xi’an Jiaotong University}, \orgaddress{\street{No. 28 Xianning West Road}, \city{Xi'an}, \postcode{710049}, \state{Shaanxi}, \country{China}}}

\affil[6]{\orgdiv{Ministry of Education Key Lab of Intelligent Networks and Network Security}, \orgname{Xi’an Jiaotong University}, \orgaddress{\street{No. 28 Xianning West Road}, \city{Xi'an}, \postcode{710049}, \state{Shaanxi}, \country{China}}}

\affil[8]{\orgdiv{Guangzhou Institute of Geography}, \orgname{Academy of Sciences}, \orgaddress{\street{No. 100 Xianlie Road}, \city{Guangzhou}, \postcode{510070}, \state{Guangdong}, \country{China}}}

\affil[7]{\orgdiv{\orgname{National Meteorological Information Center}, 
\city{Beijing}, \postcode{100044},  \country{China}}}

\affil[9]{\orgdiv{MDX Research Center for Element Strategy}, \orgname{Institute of Integrated Research, Institute of Science Tokyo}, \orgaddress{\street{Midori-ku}, \city{Yokohama}, \postcode{226-8503}, \country{Japan}}}

\affil[10]{\orgdiv{School of Physics and Information Technology}, \orgname{Shaanxi Normal University}, \orgaddress{\street{No. 620 West Chang'an Avenue}, \city{Xi'an}, \postcode{710119}, \state{Shaanxi}, \country{China}}}

\affil[11]{\orgdiv{School of Aeronautics}, \orgname{Northwestern Polytechnical University}, \orgaddress{\street{No. 127 West Youyi Road}, \city{Xi'an}, \postcode{710072}, \state{Shaanxi}, \country{China}}}

\affil[12]{\orgname{AI for Science Institute}, \orgaddress{\city{Beijing}, \postcode{100080}, \country{China}}}

\affil[13]{\orgname{DP Technology}, \orgaddress{\city{Beijing}, \postcode{100080}, \country{China}}}

\affil[14]{\orgdiv{Academy for Advanced Interdisciplinary Studies}, \orgname{Peking University} \orgaddress{\city{Beijing}, \postcode{100871}, \country{China}}}

\affil[15]{\orgdiv{School of Mathematics and Statistics}, \orgname{Xi’an Jiaotong University}, \orgaddress{\street{No. 28 Xianning West Road}, \city{Xi'an}, \postcode{710049}, \state{Shaanxi}, \country{China}}}

\abstract{We uncover a phenomenon largely overlooked by the scientific community utilizing AI: neural networks exhibit high susceptibility to minute perturbations, resulting in significant deviations in their outputs. Through an analysis of five diverse application areas—weather forecasting, chemical energy and force calculations, fluid dynamics, quantum chromodynamics, and wireless communication—we demonstrate that this vulnerability is a broad and general characteristic of AI systems. This revelation exposes a hidden risk in relying on neural networks for essential scientific computations, calling further studies on their reliability and security.}




\maketitle

\section{Introduction}\label{sec1}

\begin{figure}
\begin{centering}
\includegraphics[width=0.75\textwidth]{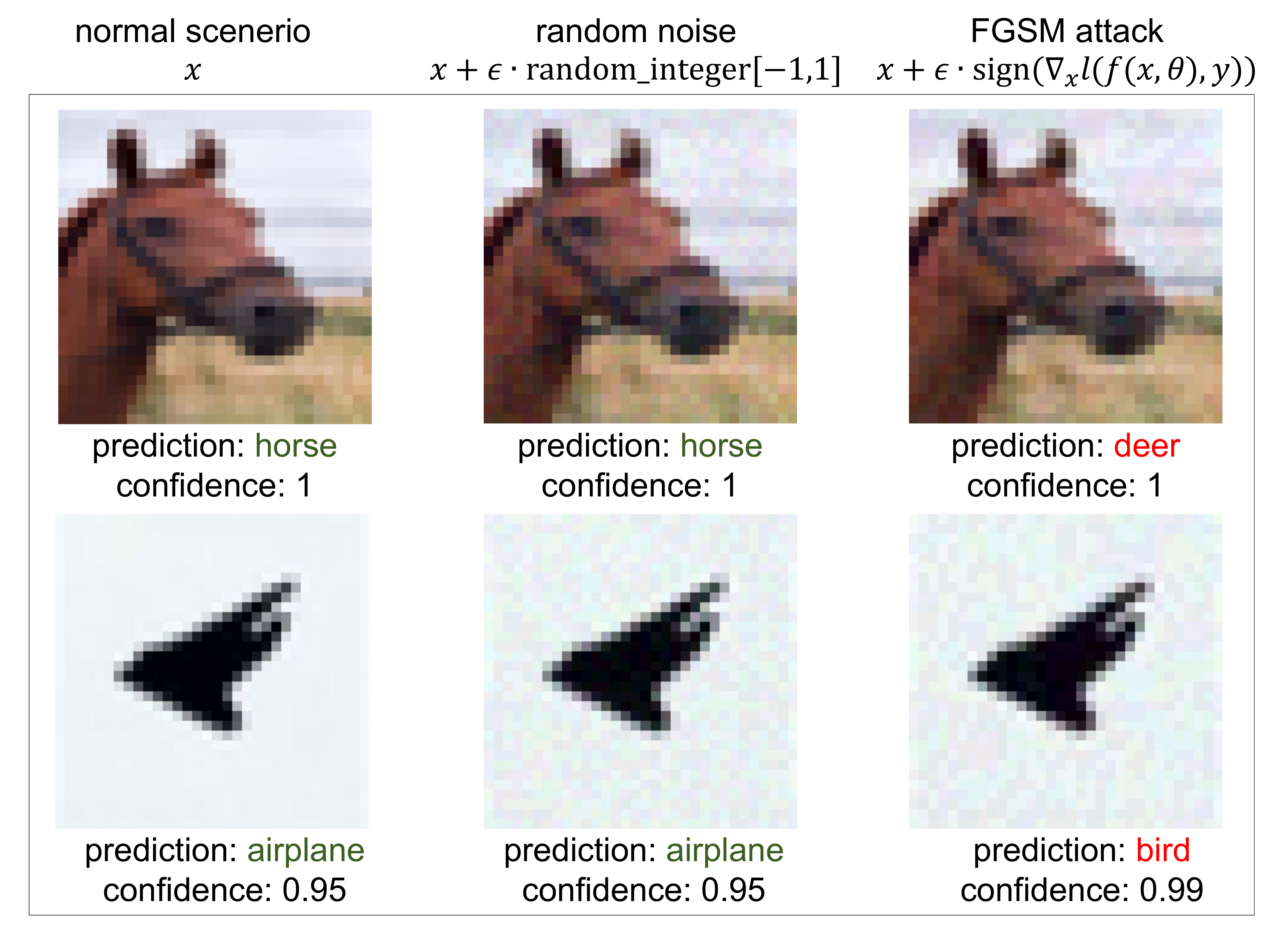}
\par\end{centering}
\caption{\textcolor{black}{Accurate neural network fails to predict the correct label with small non-random perturbations. The trained neural network is evaluated under three different conditions: normal scenario, with random noise, and under an attack using the Fast Gradient Sign Method (FGSM) \cite{goodfellow2014explaining, kurakin2016adversarial}. In the normal scenario, the network is tested using the original test dataset. For the random noise scenario, the test dataset images are modified by adding random noise with pixel values of either -$\epsilon$ or $\epsilon$, where $\epsilon=0.1$. In the FGSM attack scenario we perturb the input images by adding a small, calculated noise in the direction of the gradient of the loss function. Specifically, the transformation applied is:
$x \rightarrow x + \epsilon \cdot \text{sign}(\nabla_{x} l(f(x, \theta), y))$,
where $x$ represents the input images. It is important to note that $y$ is the true label which corresponds to an untargeted attack, where the neural network's prediction would deviate away from the true label. $l(f(x, \theta), y)$ is the loss function, and $\theta$ denotes the weights of the trained network. The trained Googlenet model \cite{szegedy2015going}, when evaluated on the CIFAR-10 dataset \cite{Alex2009}, achieves an accuracy of 89\% on both the original and random noise scenarios but only 18\% accuracy on the images subjected to the FGSM attack.}}
\label{fig:FGSM-attack-reduces}
\end{figure}

Artificial Intelligence (AI) has become a transformative tool in scientific research, driving breakthroughs across numerous disciplines \cite{jumper2021highly, senior2020improved, radovic2018machine, guest2018deep, reichstein2019deep, schultz2021can, rasp2020weatherbench}. Despite these achievements, neural networks, which form the backbone of many AI systems, exhibit significant vulnerabilities. One of the most concerning issues is their susceptibility to adversarial attacks \cite{9da342810d2f4fee9f6ada111a7891d1, goodfellow2014explaining, REN2020346, kurakin2016adversarial}. These attacks involve making small, often imperceptible changes to the input data, causing AI systems to make incorrect predictions (Fig. \ref{fig:FGSM-attack-reduces}), highlighting a critical weakness: AI systems can fail under minimal perturbations - a phenomenon completely unseen in classical methods.

The impact of adversarial attacks has been extensively studied in the context of image classification \cite{szegedy2013intriguing, papernot2016limitations, doi:10.34133/remotesensing.0219}. However, this vulnerability has not received widespread attention in other scientific domains. In scientific research, where stability and controllability of AI performance are essential, the exposure of such vulnerabilities is particularly concerning.
\textcolor{black}{Here, we investigate the vulnerability of neural networks to adversarial attacks across five scientific application areas, spanning a broader range of AI applications in scientific research: weather forecasting in climate science (data-driven sequential predictions) \cite{rasp2018deep, dueben2018challenges}, chemical energy and force calculation (data-augmented chemistry-driven tasks) \cite{schutt2017quantum, gilmer2017neural}, parametric problems in fluid dynamics (physics-informed neural networks, PINNs) \cite{ling2016reynolds, brunton2020machine}, prediction of equation of state in quantum chromodynamics (data-augmented physics-driven tasks) \cite{baldi2014searching, urban2018deep}, and wireless communication in signal processing (experiment-augmented deep reinforcement learning) \cite{jiang2017machine, zhang2019deep}.} 
By examining these diverse examples, we aim to highlight the widespread nature of neural network vulnerabilities, extending the concern beyond image processing to a broader scientific audience.

\section{Results}\label{sec2}

\subsection{Weather forecasting}

\begin{figure}
\begin{centering}
\includegraphics[width=0.95\textwidth]{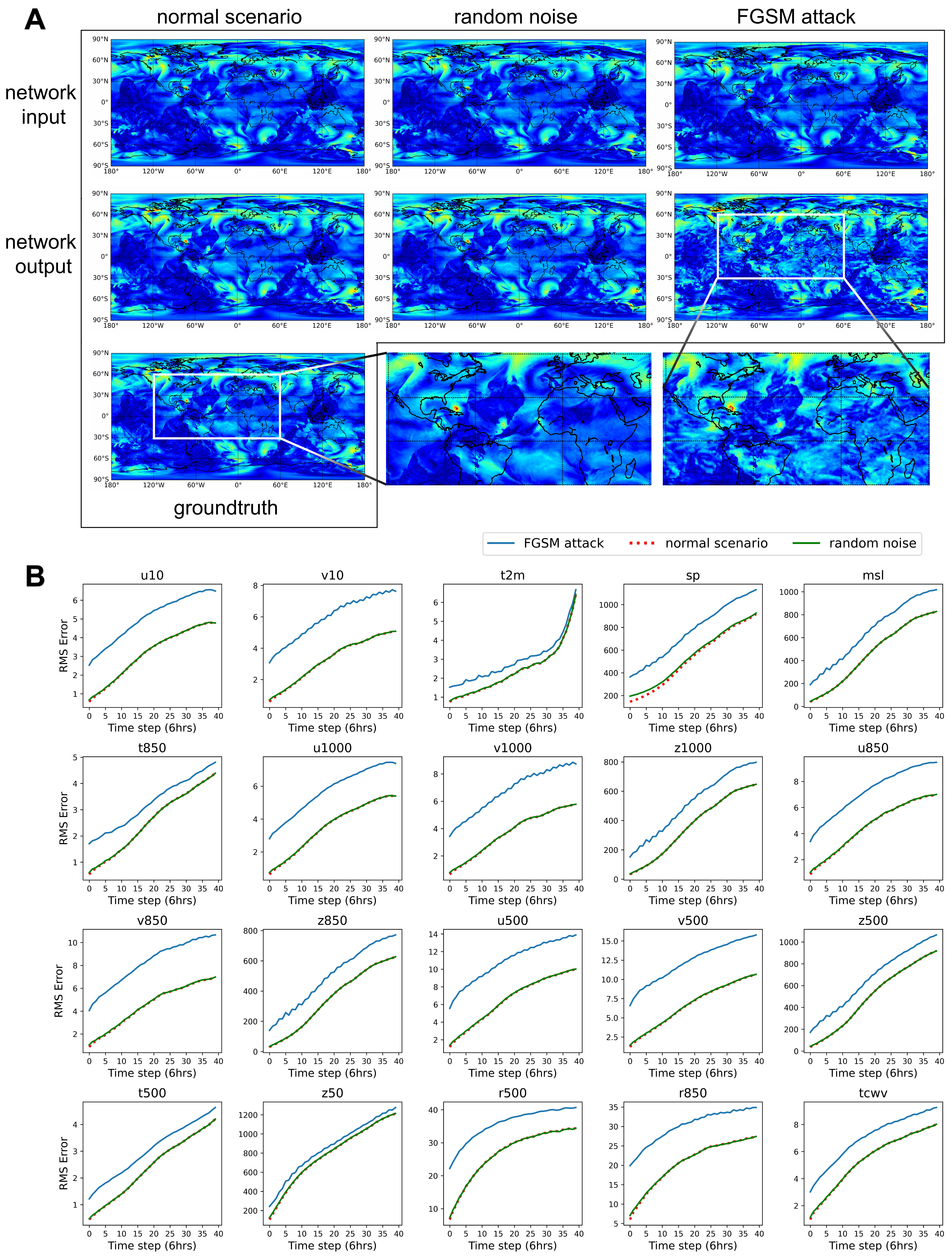}
\par\end{centering}
\caption{\textbf{Testing results of the FourCastNet under perturbations}. Upper panel \textbf{A} shows the wind speed distribution at a selected time step. The input (first row) is perturbed using the FGSM attack with an epsilon value of 0.05 (roughly changing the input by 5\%), as well as random noise values of -0.05 and 0.05. The corresponding predictions are displayed in the second row, with the ground-truth for this time step shown in the first column of the third row. A specific region in the ground-truth and the FGSM attack prediction is magnified for detailed comparison. Panel \textbf{B} compares the predictive root mean square error (RMSE) between the normal scenario, input perturbed by the FGSM attack, and random noise. The 20 atmospheric variables include surface-level variables such as U10, V10, T2M, SP, and MSLP; variables at 1000 hPa including U, V, and Z; variables at 850 hPa and 500 hPa including T, U, V, Z, and RH; a variable at 50 hPa, which is Z; and an integrated variable, TCWV.}
\label{fig:Fourcastnet-results}
\end{figure}

FourCastNet \cite{10.1145/3592979.3593412}, an abbreviation for Fourier ForeCasting Neural Network, is a state-of-the-art global data-driven weather forecasting model that leverages the Adaptive Fourier Neural Operator (AFNO) architecture. The AFNO integrates the Fourier Neural Operator approach with a Vision Transformer backbone.  FourCastNet provides delivers notably accurate short- to medium-range global predictions at a 0.25° resolution, capable of producing a week-long forecast in less than 2 seconds- substantially faster than traditional numerical weather prediction models such as the ECMWF (European Center for Medium-Range Weather Forecasts) integrated forecasting system, while achieving comparable or even superior accuracy.

In our study, we employ a pretrained model from the FourCastNet GitHub repository to investigate its susceptibility in predicting 20 atmospheric variables. The model is trained on a subset of the fifth generation ECMWF reanalysis (ERA5) dataset \cite{hersbach2018era5}, which encompasses hourly estimates of various atmospheric variables. We then introduce slight perturbations (roughly 5\% deviation) to the input data using the FGSM attack and random noise, both of which are imperceptible (with the magnitude of perturbations equivalent for both methods, differing only in direction). The results are illustrated in Fig. \ref{fig:Fourcastnet-results}.

In Fig. \ref{fig:Fourcastnet-results} \textbf{A}, we observe numerous undesirable wind distribution patterns in the FGSM attack predictions compared to the ground-truth. This discrepancy is absent in the normal (without any perturbations in the input) and random noise scenarios, indicating that only specific perturbations, such as those introduced by the gradient-based FGSM attack, significantly degrade the network's performance. Similarly, Fig. \ref{fig:Fourcastnet-results} \textbf{B} shows that the FGSM attack results in a higher Root Mean Square (RMS) error compared to both the normal and random noise conditions, highlighting FourCastNet's inability to accurately predict certain patterns.

The vulnerability of FourCastNet holds significant implication for climate science. Perturbations in input data - arsing from factors such as intentional interference with satellite sensing, non-random atmospheric disturbances, or inherent data correlations - may partially or wholly manifest as gradient-based perturbations. The vulnerability contrasts sharply with classical algorithms, where minor input alterations typically yield acceptable output variations. In neural network, however, even slight input changes can lead to highly unpredictable outputs, presenting a critical challenge to the reliability of applications. Such perturbations can lead to substantial deviations in the model's predictions, potentially causing irreversible consequences on decision-making.

\subsection{Chemical energy and force calculation}

\begin{figure}[h]
\centering
\includegraphics[width=1\textwidth]{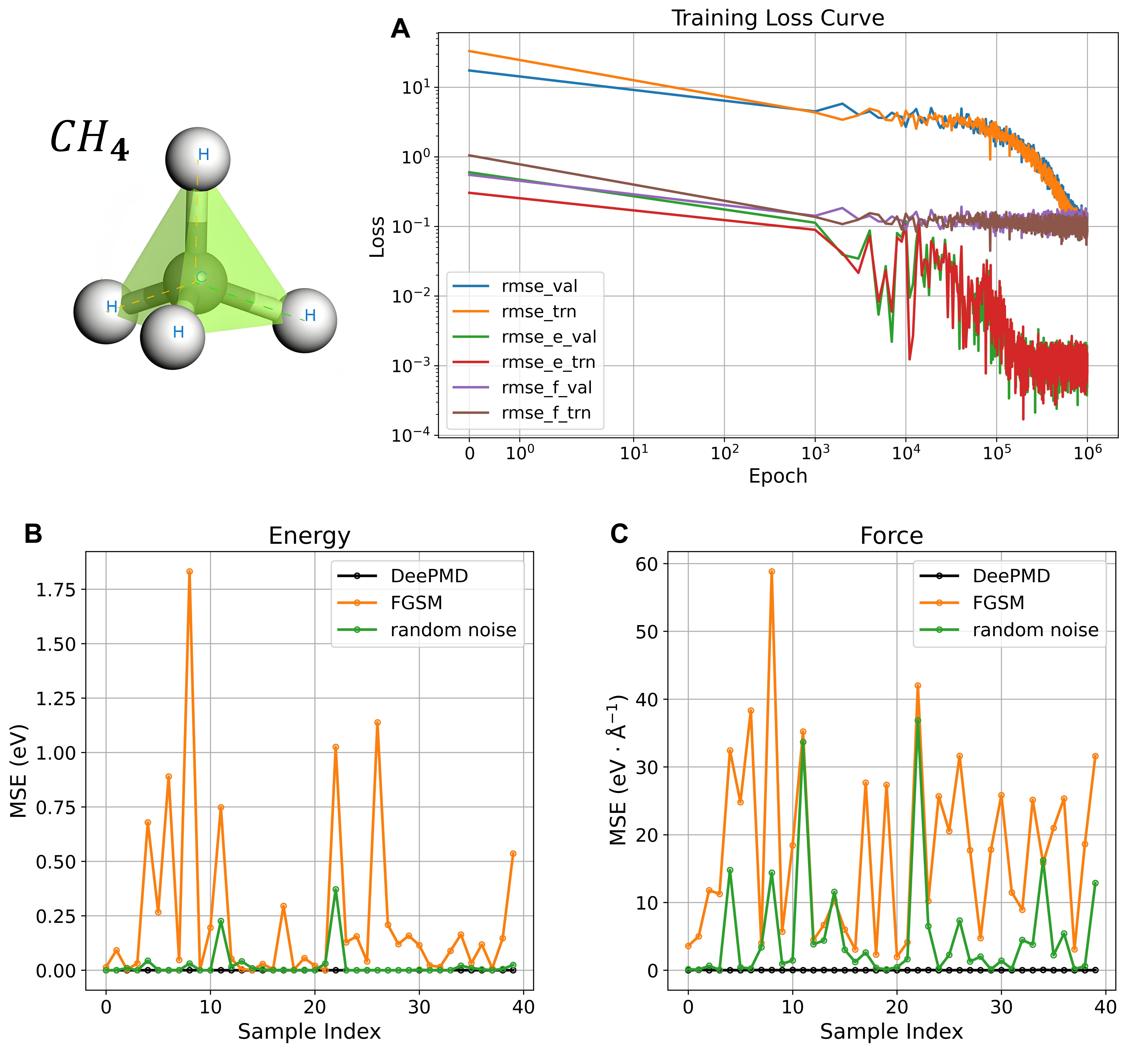}
\caption{\textbf{Training and test results using the DeePMD-kit in calculating the energy and force of the methane molecule}. \textbf{A}, the training loss curve of the network. \text{``rmse\_val"}, \text{``rmse\_trn"}, \text{``rmse\_e\_val"}, \text{``rmse\_e\_trn"}, \text{``rmse\_f\_val"}, \text{``rmse\_f\_trn"}, refer to the validation loss, training loss, RMS validation error of energy, RMS training error of energy, RMS validation error of force, RMS training error of force. \textbf{B} and \textbf{C}, comparison of MSE between DeePMD-kit predictions and VASP calculations for methane molecules in the test phase. The test is performed on 40 different frames with varying coordinates and box dimensions. For each frame, the coordinates are slightly perturbed by noise values of either -0.05 or 0.05, applied using the FGSM attack (orange) or random noise (green).}
\label{deepmd-results}
\end{figure}

DeePMD-kit \cite{WANG2018178, 10.1063/5.0155600} is a deep learning-based tool designed to simulate the behavior of molecules and materials with high accuracy and efficiency. It employs deep neural networks (DNN) to model the potential energy surface of a system by decomposing the total energy into contributions from individual atoms and their local environments. The DNN maps atomic environment descriptors to atomic energies with high precision through several innovative features: symmetry preservation, efficient training, and molecular dynamics package integration. This approach ensures that the model respects physical symmetries such as translation and rotation. Trained with high-accuracy quantum mechanical data, DeePMD-kit achieves accuracy comparable to advanced methods like density functional theory but at a fraction of the computational cost, enabling simulations of larger systems and longer timescales.

We use DeePMD-kit to calculate the energy and force of methane molecules and demonstrate the vulnerability of the trained model to perturbations. The neural network was trained using 200 data frames generated by VASP (Vienna Ab-initio Simulation Package \cite{Kresse1993, Kresse1996a}), with 160 frames for training, 40 frames for validation, and another 40 frames for testing. After training for 1,000,000 epochs, the MSE loss of the training and validation data stabilized (Fig. \ref{deepmd-results} \textbf{A}), allowing us to proceed to the test phase.

During testing, the input coordinates of the methane molecules were perturbed using the FGSM attack and random noise, both introduce about 1\% variation around their original coordinates. The perturbed coordinates were also provided to VASP to obtain corresponding energy and force values in order to obtain the relevant MSE. The test results are shown in Fig. \ref{deepmd-results} \textbf{B} and \textbf{C}. Given the slight perturbations in the coordinates, the predictions from the neural network should not significantly deviate from the VASP calculations. However, we observe that the predicted energy and force from the well-trained network both suffer from severe deviations, especially when the FGSM attack is performed, indicating its vulnerability to specific perturbation patterns. 

\textcolor{black}{The vulnerability of DeePMD reveals a fact that the trained network is unstable under specific patterns. In practical applications, the coordinates of methane molecules input to DeePMD may experience a deviation of about 1\%. If the displacement of each atom coincides with the direction of an FGSM attack, the prediction results of DeePMD will become unreliable. Therefore, it is necessary to implement targeted corrections to address the vulnerability of neural networks. To this extent, we need more powerful network structures containing more human experience \cite{zhang2024pretraining, zhang2024dpa2}.}

\subsection{Parametric problems in fluid dynamics}

\begin{figure}
\begin{centering}
\includegraphics[width=1\textwidth]{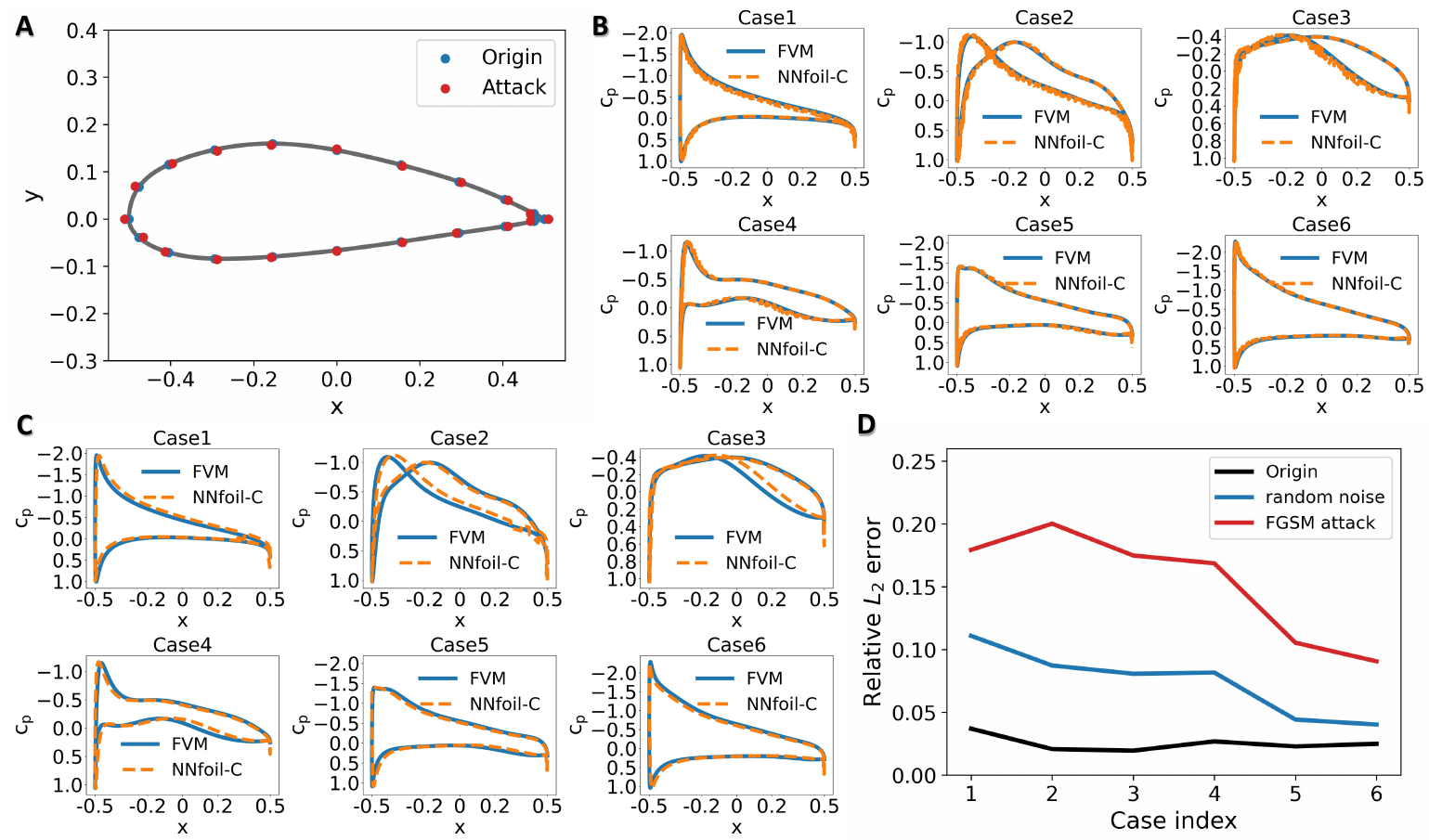}
\par\end{centering}
\caption{\textcolor{black}{\textbf{NNfoil-C predictions under different perturbations}. Panel \textbf{A} represents the perturbations added to spatial coordinates of airfoil surface sampling points (illustrative only; actual number of sampling points is ten times that shown). \textbf{B} and \textbf{C} compare the airfoil surface pressure coefficient distributions output by NNfoil-C after adding random noise and FGSM attacks to the six cases used in the literature \cite{cao2024solving} with the reference solutions obtained through the finite volume method (FVM). \textbf{D} gives relative L2 error between the airfoil surface pressure coefficient distributions of the six cases output by the original NNfoil-C and NNfoil-C after two types of attacks and the reference solution.}}
\label{fig:inverse_fluid}
\end{figure}

Flow simulation is a crucial branch of fluid mechanics that involves using computers to solve fluid governing equations,  which has extensive applications in fields such as aircraft design, weather forecasting, and natural gas transportation. In engineering practice, we need to evaluate flow fields under numerous different flow conditions or shapes. Traditional numerical methods can only solve a single flow condition or shape at a time, resulting in high computational costs. In contrast, PINNs can efficiently solve a series of similar problems simultaneously, especially in high-dimensional parametric problems. A notable example is NNfoil-C \cite{cao2024solving}, an unsupervised complete state-space airfoil flow solution framework. By extending the conventional PINNs input, NNfoil-C constructs a high-dimensional complete state-space, including spatial coordinates, Mach number, angle of attack, and 12 class shape transformation parameters controlling the airfoil shape. With a single training session, NNfoil-C can obtain flow fields for all flow conditions and shapes within the complete state-space in just 18.8 hours, whereas traditional numerical methods require a time that is several orders of magnitude longer. 

In our research, we investigate the vulnerability of NNfoil-C to perturbations in spatial coordinates. Specifically, we attack only the spatial coordinates in the model input, keeping flow condition parameters (Mach number, angle of attack) and the shape parameters unchanged, as minor changes in the latter two may significantly alter the flow field in real scenarios. We introduce a 3\% perturbation in the spatial coordinates (as shown in Fig. \ref{fig:inverse_fluid} \textbf{A}) on the airfoil surface using random noise and FGSM attacks. These small changes in coordinates should not significantly alter the network's predictions in principle. It can be observed that adding random noise to the input only causes slight fluctuations in the output, with the overall trend of the curve closely matching the reference solution (Fig. \ref{fig:inverse_fluid} \textbf{B}). However, adding FGSM attacks to the input results in a noticeable shift in the output, especially in cases 1-4 (Fig. \ref{fig:inverse_fluid} \textbf{C}). To visualize the impact of FGSM on the network more clearly, Fig. \ref{fig:inverse_fluid} \textbf{D} shows the relative L2 error between the airfoil surface pressure coefficient distributions of the six cases after perturbation and the reference solution. The error after adding random noise is generally below 10\%, not significantly affecting the network's performance, while the error after FGSM attacks is almost always above 10\%, reaching up to 20\% in some cases. This deviation is entirely unacceptable in subsequent airfoil force calculations.

These results reveal potential challenges in neural network-based PDE (Partial Differential Equation) solution frameworks. In practical applications, deviations in surface coordinate measurements from experiments are normal. If the deviation direction strongly aligns with the FGSM attack direction, or if a significant portion aligns with FGSM, the model's output becomes unreliable. Additionally, the results indicate that the high-dimensional space fitted by PINNs is not as smooth as expected. If targeted attack coordinates can be designed for any set of sampled coordinates, it suggests that a considerable portion of the solution space in PINNs significantly deviates from the analytical solution. The phenomenon underscores the necessity of conducting robustness research on PINNs, a direction currently unexplored in the field.

\subsection{Quantum Chromodynamics Equation of State}

\begin{figure}[h]
\centering
\includegraphics[width=1\textwidth]{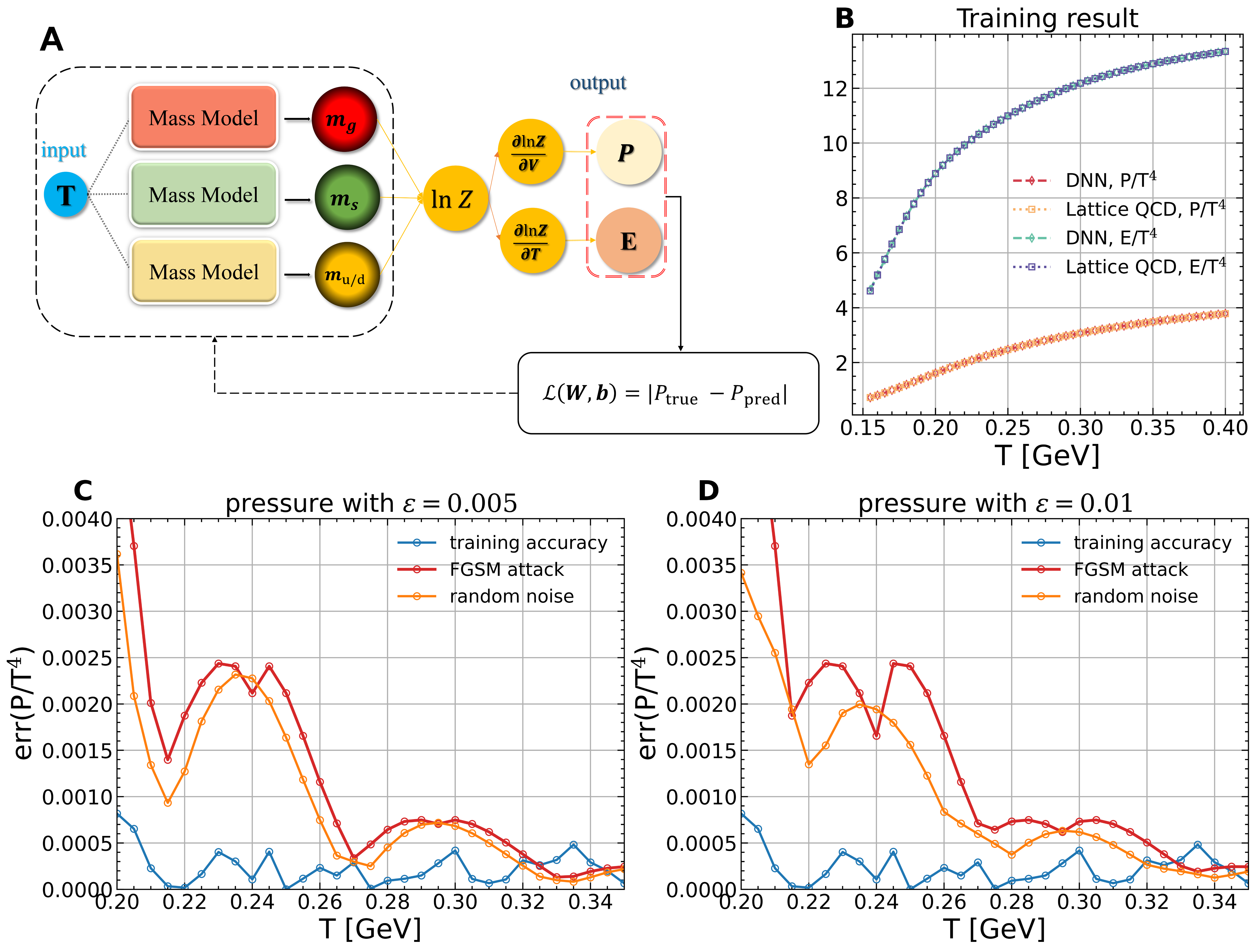}
\caption{\textbf{Deep-learning quasi-particle model developed to reproduce the QCD equation of state}. \textbf{A}, the framework of the neural network, where the input is the one-dimensional temperature and the output is the corresponding quasi-particle mass, contributing to the partition function. \textbf{B}, the pressure and energy density as functions of temperature using the network model as well as the lattice QCD calculations. \textbf{C} and \textbf{D}, the test errors of the predicted pressure and lattice QCD calculations as a function of temperature under 3 different conditions: (1) no perturbations are involved (training accuracy), (2) the temperatures are slightly perturbed with values of either $\epsilon$ or -$\epsilon$, using the FGSM attack (red), and (3) the random noise (orange), for $\epsilon=0.005$ and $\epsilon=0.01$.
}
\label{fig:QCD-result}
\end{figure}

Quantum Chromodynamics (QCD) is a theory that describes the strong interaction between quarks and gluons, which is one of the four fundamental forces. Lattice QCD, leveraging supercomputers, derives the properties of nuclear matter under extreme conditions through first-principle calculations and expresses them via an equation of state (EoS). We have developed an effective model for strongly coupled quark-gluon plasma (Deep Learning Quasi Particle - DLQP model) using a physics-based neural network, which treats the plasma as a quasi-parton gas with weak interactions. This model allows for the convenient reproduction of the QCD EoS on personal laptops using simple statistical formulas \cite{Li:2022ozl}. The model employs a DNN to represent the temperature-dependent masses of quasi-quarks (u/d and s quarks) and quasi-gluons as shown in Figure \ref{fig:QCD-result} \textbf{A}. Figure \ref{fig:QCD-result} \textbf{B} showcases the training outcomes using the DLQP model, which accurately describes the EoS computed by lattice QCD. As illustrated, despite employing pressure as the sole training objective, the model excels in predicting both pressure and energy density, closely matching the results from lattice QCD calculations.

To illustrate the model's vulnerability, we introduced FGSM attacks and random noise into the input temperature with perturbation values of either $\epsilon$ or $-\epsilon$. The results are shown in Fig. \ref{fig:QCD-result} \textbf{C} and \textbf{D}.  For most temperature values within the range [0.20, 0.34] GeV, the errors induced by FGSM attacks are significantly larger than those from random perturbations. This is expected, as the FGSM attack specifically targets the direction in which error increases most rapidly. However, there exist certain temperatures where random noise introduces larger errors than FGSM attacks. For instance, at $T=0.24$ GeV, the error induced by random noise is greater. \textcolor{black}{This interesting observation reveals a feature of the local landscape of the loss as a function of the input variable $T$: the gradient is the largest along one direction in the short range but becomes larger along another direction in a longer range as shown in the blue line at $T=0.24$ GeV. Therefore, if the amplitude of the perturbation is larger than the short range scale, the random noise will introduce larger errors than FGSM attacks. Generally, however, the FGSM attacks will introduce larger errors than random noise. }

The vulnerability analysis for one-dimensional input in this problem is essential for understanding the loss landscape in high-dimensional space. A similar phenomenon, where random noise introduces higher error, has been observed in parallel studies within this paper. The one-dimensional analysis revealed that the error difference between the FGSM attack and random noise is determined by the loss landscape for various input samples. In high-dimensional space, the model exhibits robustness against smooth loss landscapes; in such cases, the FGSM attack introduces errors comparable to those caused by random noise. Conversely, for rapidly fluctuating loss landscapes, a minor perturbation induced by the FGSM attack results in significantly larger errors than those caused by random noise. 

In the quasi-parton model of QCD equation of state, part of the uncertainty stems from Lattice QCD data, while another portion arises from the neural network itself. Investigating the sensitivity of errors introduced by the FGSM attack and random noise aids in identifying the uncertainty of the neural network model. This exploration may guide the selection of neural network structures with smoother loss landscapes, potentially enhancing model robustness and accuracy.

%of the QCD EoS. Due to the limited data provided by lattice QCD, continuous interpolation and extrapolation are often required. Determining whether anomalous behavior in the QCD EoS is due to systematic errors or fluctuations becomes particularly important. In future research and analysis, we can use the FGSM method to quantitatively analyze the uncertainties in the EoS, thereby avoiding erroneous physical conclusions. Additionally, it necessitates the use of more physical constraints to enhance the robustness of the neural network.

% The variations in prediction error are significant in the field of QCD. Since the problem is essentially a one-dimensional curve-fitting task, we should expect the fitted curve to be smooth and stable. This phenomenon reveals a critical feature of DNNs that requires further discussion to enhance reliability, especially in areas like QCD. Notably, the added "random noise" also introduces errors of similar magnitude to the FGSM attack, suggesting that as long as a sufficient portion of the perturbation coincides with the FGSM attack, the network can fail in predicting the correct labels. For a one-dimensional task, there is a significant chance for random noise to coincide with the FGSM attack.

\subsection{Wireless transmission in communication}
\begin{figure}
    \centering
    \includegraphics[width=0.95\linewidth]{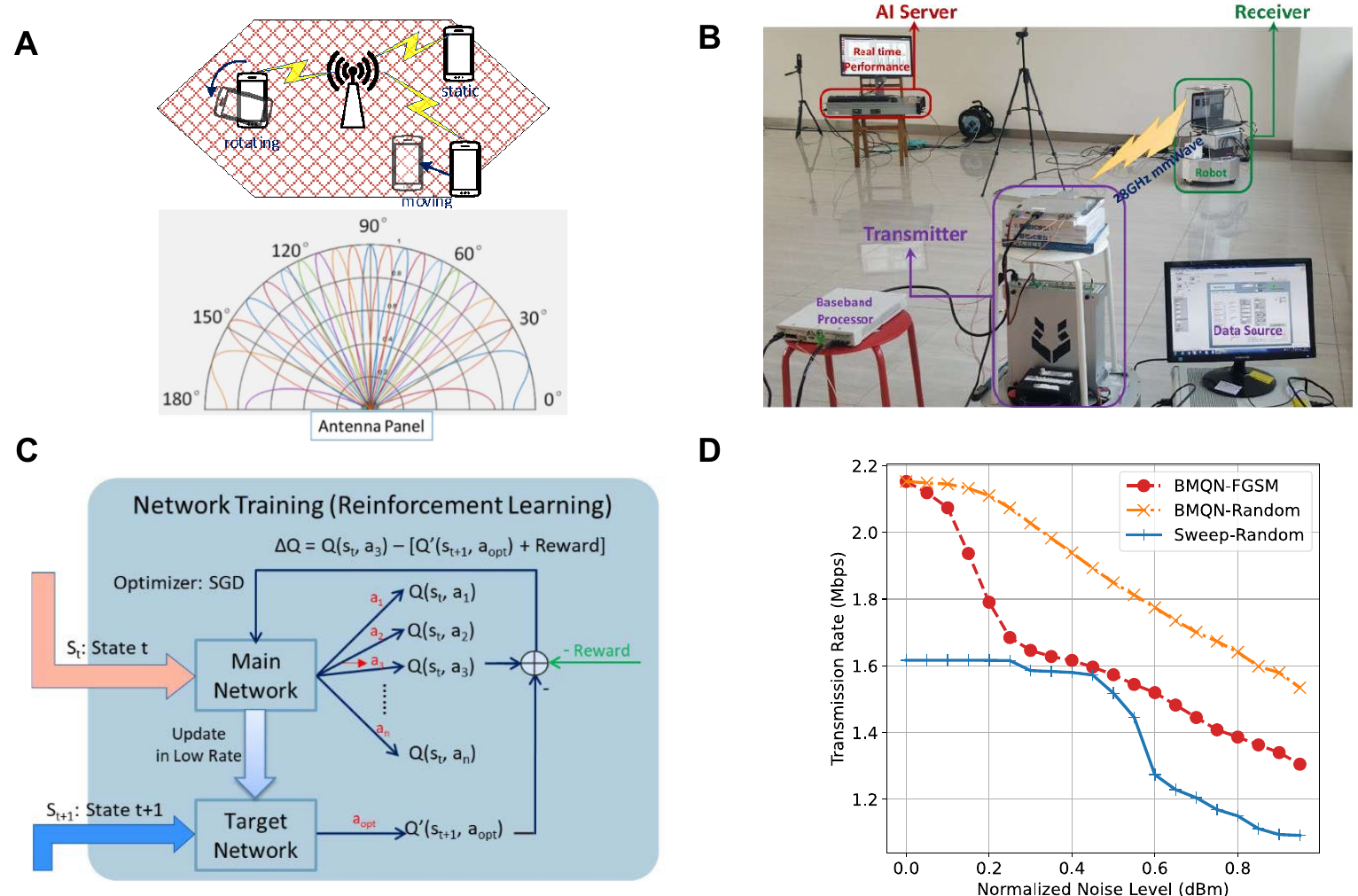}
    \caption{\textbf{Experimental setup and results of AI-based Beam Management Q-Network (BMQN) under interference scenarios}. Panel \textbf{A} demonstrates the scenario where the UE (mobile phone) is either moving or rotating, causing unstable communications. The antenna has 16 beam directions for communication. Panel \textbf{B} shows the hardware setup. Panel \textbf{C} illustrates the off-line reinforcement learning framework used to train BMQN, where the main network outputs the policy based on the temporal sequence of beam measurements, and the target network provides stable updates to improve prediction accuracy. Panel \textbf{D} provides a comparative analysis of communication rates for the BMQN and traditional sweep methods under varying perturbation amplitudes. The graphical representation illustrates the performance of three distinct curves: BMQN's response to FGSM noise attacks (red), BMQN's performance under stochastic noise (yellow), and the traditional sweep method's communication rate with stochastic noise (blue), where dBm = 0 representing the baseline communication rates without noise interference.}
    \label{fig-bm}
\end{figure}

Here, we introduce the Beam Management Q-Network (BMQN), an AI-based approach that employs reinforcement learning to predict and select the optimal beam in real time based on historical measurements. Theoretically, the optimal beam selection in the UE moving scenario (Fig. \ref{fig-bm} \textbf{A}) requires simultaneously measuring all possible beams — typically 16 beams — and then making a choice based on the feedback. However, this process is impractical with current hardware limitations. The BMQN model (Fig. \ref{fig-bm} \textbf{C}) circumvents this by only needing to communicate with three fixed  beams on UE to determine which one is better, thus approximating near-optimal performance within the constraints of actual 5G systems. Under normal conditions, our AI-based BMQN demonstrates high transmission rates comparable to those achieved through optimal beam selection. \textcolor{black}{This effectiveness is shown in Fig. \ref{fig-bm} \textbf{D} with dBm=0, where the BMQN’s performance outperforms the reference traditional sweep-based algorithm \cite{bai2016analysis}, underscoring the practical potential of reinforcement learning for beam management in 5G applications. }

However, such AI model suffers from potential vulnerability that are unpredictable and severe. To evaluate the resilience of the AI-driven beam selection algorithm, we introduce two types of interference in the test phase: a focused FGSM-based adversarial attack and random noise. The FGSM attack injects a targeted interference signal that slightly shifts the channel state measurements perceived by the UE - thus altering the input data fed to the AI model. By contrast, the random noise represents similar interference with equivalent amplitude. 

As shown in Fig. \ref{fig-bm} \textbf{D}, the FGSM attack significantly impacts the performance of the BMQN, where a mere 0.2 dBm (about 5\% deviation) of normalized noise reduces the communication rate from 2.2 Mbps to 1.6 Mbps—a decline of over 27\%. At this noise level, the BMQN-based beam selection method's performance deteriorates to match the traditional sweep-based method's performance. If such a small disturbance can reduce the advantages of AI to a level close to that of classical algorithms, then the numerous costs incurred in training AI lose their value.

These findings reveal the vulnerability of intelligent wireless networks to targeted adversarial attacks, suggesting the need to improve wireless network design with defensive strategies that can mitigate the impact of intentionally crafted noise interference, particularly in emerging 5G millimeter-wave communication technologies.

\section{Discussion}

\subsection{Vulnerability of High-Precision Neural Networks in Scientific Research}

\textbf{Trustworthiness of Solutions.} The trustworthiness of solutions provided by neural networks is a practical concern. For instance, in examples like the ``Quantum Chromodynamics Equation of State" and ``Parametric Problems in Fluid Dynamics" discussed in our article, the inputs are continuous physical quantities such as temperatures or spatial coordinates. If the network has indeed learned all the features of the solution space, any small changes in the inputs should not result in unacceptable deviations from the ground-truth. However, the vulnerability observed indicates that neural networks tend to minimize the loss function at training points but leave large loss values in between (more often between features and concepts, as discussed in Ref. \cite{zhang2024exploring}), leading to a non-smooth loss landscape. This raises concerns for computational sciences, where inputs are continuous physical variables, as such a non-smooth loss landscape suggests that the solution may not align with the underlying physical laws.

\textbf{Impact of Random Noise.} In most real-world scenarios, inputs may not encounter the exact gradient perturbations, thus the impact of random noise on neural network performance is generally weaker. However, in certain cases, such as low-dimensional inputs (e.g., QCD, wireless communication, and small molecule chemical potentials), random noise has a higher probability of coinciding with gradient-based attacks. This overlap reveals potential threats when these algorithms are used on a large scale, indicating that even in the absence of deliberate attacks, the inherent noise in data can pose risks to the accuracy and reliability of AI models.

\textbf{Susceptibility to intentional Attacks.} Many AI applications are critical to national infrastructure and public welfare, such as the above examples of wireless communication and weather forecasting. If these AI systems can be easily attacked, there will be potential risks in exploiting AI systems to achieve specific objectives, thereby reducing the reliability and security of essential services. Moreover, these attacks are often imperceptible, making them difficult to detect, which amplifies the risk and potential impact of such activities.

\subsection{Inherent Nature of Neural Network Vulnerability}

A more critical concern is whether the observed vulnerability stems from insufficient data and weaker network designs or is an inherent characteristic of neural networks. This question remains unresolved.

The behavior of the loss landscape in terms of the input/feature space provides a perspective to understand the vulnerability phenomenon. When training a neural network, the features or concepts of inputs are gradually extracted. These features share sharp boundaries measured by the loss function in terms of the input $x$ (or features) \cite{mikolov2013distributed, pennington2014glove, zhang2024exploring}. Near these sharp boundaries, a small disturbance along the gradient (i.e., the gradient of the loss function with respect to the input or feature) tends to move one concept to another, leading to the network being attacked. Therefore, the vulnerability of neural networks can be seen as a natural result of training and being accurate. 

Such vulnerabilities can be further understood and quantified by the "uncertainty principle of neural networks" \cite{10.1093/nsr/nwae141}, where the network input and gradient attack are recognized as two conjugate variables, like position ($x$) and momentum ($p \sim \nabla_x \phi(x)$, where $\phi(x)$ is the wave function) in quantum mechanics. In quantum mechanics, one cannot measure the position and momentum of a particle simultaneously to an arbitrarily high accuracy. Following a similar analogy, the input and gradient attacks cannot be resolved accurately by the trained network, leading to vulnerability.

\subsection{Possible Directions in Making Robust Networks}

If network vulnerability is a natural outcome of training and a manifestation of sharp boundaries between concepts, avoiding such training or smoothing these boundaries could be promising directions. One such candidate is the Randomized Neural Network (RNN) \cite{SHANG2023107518, SUN2024115830}, where the network weights are not fully trained, thus avoiding sharp boundaries between concepts.

To see the effectiveness of RNN, we solve the two-dimensional Poisson equation using a DNN and RNN separately. Then we attack the two networks using FGSM with the same perturbation magnitudes. The results are shown in Tab. \ref{tab:rnn_dnn_attack_performance}. We see that RNN is more robust than DNN, and the difference is more prominent under relative strong attacks.

\begin{table}[h!]
\centering
\begin{tabular}{|c|c|c|c|c|c|c|c|}
\hline
\textbf{Attack Amplitude $\epsilon$} & 0.01 & 0.05 & 0.1 & 0.15 & 0.2 & 0.25 & 0.3 \\ \hline
\textbf{RNN Rel MSE}                & 0.442 & 0.539 & 0.087 & 0.053 & 0.273 & 0.234 & 0.379 \\ \hline
\textbf{DNN Rel MSE}                & 0.390 & 1.459 & 0.882 & 2.092 & 3.109 & 3.304 & 4.061 \\ \hline
\end{tabular}
\caption{Performance of RNN and DNN under attacks of the same amplitude. Rel MSE is calculated as $|\text{Attacked MSE} - \text{Original MSE}|$/$\text{Original MSE.}$}
\label{tab:rnn_dnn_attack_performance}
\end{table}

\section{Conclusion}

As AI continues to be integrated into various fields, the fragility of these systems will likely become more apparent, potentially influencing their reliability and effectiveness. Meanwhile, our study also verifies that this phenomenon is not isolated and suggests that the vulnerability of AI systems is an inherent property, which can be quantified by the uncertainty principle of neural networks \cite{10.1093/nsr/nwae141}.

In conclusion, we highlight a critical aspect that has been largely overlooked in the community: the robustness of scientific AI models. Our findings reveal that neural networks, despite their high precision and efficiency, exhibit significant vulnerabilities to minute perturbations. These vulnerabilities can lead to substantial deviations in outputs, posing risks to the reliability and security of AI-driven solutions in various scientific domains. However, our work is by no means to undermine the value of AI in scientific research but to raise awareness about the importance of robustness in AI models and to encourage the development of more resilient and trustworthy AI systems.

\section{Methods}\label{sec3}

\subsection{Weather Forecasting}

\subsubsection{Model Overview}
FourCastNet employs an Adaptive Fourier Neural Operator (AFNO) network to forecast the dynamics of 20 atmospheric variables at future time steps, based on an initial condition drawn from the ERA5 dataset. The model iteratively applies the AFNO network to generate these predictions. The variables include surface-level winds, temperature, pressure, and other essential atmospheric metrics across various pressure levels (1000 hPa, 850 hPa, 500 hPa, and 50 hPa), as well as integrated variables such as Total Column Water Vapor (TCWV).

For weather prediction, the model takes 20 different ERA5 variables, each represented on a regular latitude/longitude grid of dimension $720\times 1440$ that covers the entire globe at a specific initial time step $t$. Subsequently, the AFNO architecture forecasts these variables for a later time step $t+\Delta t$, where the original paper defines a fixed time interval $\Delta t$ of 6 hours.

\subsubsection{Inference and Perturbations}

\textbf{Inference Process.}
The inference process involves autoregressive prediction, where the model's output at each time step is fed back as input for the subsequent time step, enabling the model to forecast multiple steps ahead. In our experiments, we directly loaded the pretrained model checkpoint provided by the authors to perform inference on the 2018 data.

\textbf{Adversarial Attacks.}
To evaluate the robustness of the FourCastNet model, we implemented adversarial attacks using the FGSM and random perturbations. The FGSM attack perturbs the input data by adding a small perturbation in the direction of the gradient of the loss relative to the input. Conversely, the random perturbation attack introduces a random noise tensor of equivalent magnitude as the FGSM perturbation, but with random directions.

During testing, we selected 10 different initial conditions from different time periods in 2018, conducting inference for each to yield ten independent predictions. Perturbations were applied at each step of the autoregressive prediction, meaning that the output from each step was also modified.
The predictions for the 20 parameters were then evaluated across the 10 independent predictions, and the RMSE was calculated.

\subsection{Molecular Dynamics with DeePMD-kit}

\subsubsection{Model Overview}
DeePMD-kit is a package written in Python/C++, designed to minimize the effort required to build deep learning-based models of interatomic potential energy and force fields, and to perform molecular dynamics. This approach addresses the accuracy-versus-efficiency dilemma in molecular simulations. Applications of DeePMD-kit span from finite molecules to extended systems and from metallic systems to chemically bonded systems. For more details, please refer to the official documentation: \url{https://docs.deepmodeling.com/projects/deepmd/en/master/getting-started/quick_start.html}.

The training data utilized by DeePMD-kit comprises essential information such as atom type, simulation box, atom coordinates, atom forces, system energy, and virial. A snapshot of a molecular system that includes this data is called a frame. Multiple frames with the same number of atoms and atom types make up a system of data. For instance, a molecular dynamics trajectory can be converted into a system of data, with each time step corresponding to a frame in the system.

In our methane molecule example, 160 frames are picked as training data, and the other 40 frames are used for validation. In the test phase, we use an additional 40 frames for testing.

\subsubsection{Inference and Perturbations}

\textbf{Inference Process.}
The inference process in DeePMD-kit involves using the trained model to predict the potential energy and forces of a molecular system. For our experiments, we trained the model for 1,000,000 epochs using the command ``dp train input.json". After training, we modified the source code to disable gradient updates during inference, effectively switching the model to test mode while still allowing gradient calculations.

\textbf{Adversarial Attacks.}
To evaluate the robustness of the DeePMD model, we implemented adversarial attacks using FGSM and random perturbations. The FGSM attack perturbs the input data by adding a small perturbation in the direction of the gradient of the loss with respect to the input. The random perturbation attack adds a random noise tensor of the same magnitude as the FGSM perturbation but with random directions.

Since it is difficult to perform these attacks directly at the API level, we modified the source code in the file ``deepmd/pt/train/training.py" to incorporate these perturbations. Additionally, we changed the ``numb\_steps" parameter in the ``input.json" file from 1000000 to 1000020 to run the network in test mode for an additional 20 epochs while loading the previously trained model. This procedure is performed using the command ``dp train input.json --restart". During the attack, we applied the perturbations to the coordinates of methane molecules, with an epsilon value of 0.05 for both FGSM and random noise. The perturbed coordinates were then used to perform inference via DeePMD-kit, and the results were saved for comparison and analysis.

The true energy and forces also change when the coordinates are perturbed, thus we saved the perturbed coordinates and input them into the VASP program to obtain the corresponding true energy and forces. This step is crucial because even small changes in the coordinates of methane molecules can lead to changes in the output. By comparing the predicted values with the true values obtained from VASP, we then calculated the MSE between the predicted and true values to evaluate the impact of the perturbations.

\textbf{VASP calculation.}
All density functional theory (DFT) calculations were performed using the Vienna \textit{ab initio} simulation package (VASP) in this work \cite{Kresse1996}. The projector-augmented wave (PAW) pseudopotentials were employed to account for the interactions between valence electrons and ionic cores \cite{Kresse1999}. The Perdew-Burke-Ernzerhof (PBE) functional of the generalized gradient approximation (GGA) was used to describe electronic exchange and correlation \cite{Perdew1996}. A plane wave basis set with a cutoff of 400 eV was adapted to expand the wave functions. To obtain the accurate energy and force of CH$_4$ molecule, we put it into a cell with varied cell parameters without structural relaxation. The energy convergence criteria were set to $10^{-6}$. The $\Gamma$-centered k-point sampling of $1 \times 1 \times 1$ was done by the Monkhorst-Pack method \cite{Monkhorst1976}.

\subsection{Parametric problems in fluid dynamics}

\subsubsection{Model Overview}
NNfoil-C \cite{cao2024solving} is an unsupervised, inviscid airfoil flow parameterization framework based on PINNs. It aims to obtain all flow fields within a complete state-space, encompassing various flow conditions and airfoil shapes, through a single training session. The method first employs a mesh transformation approach to convert the flow from physical space to computational space, mitigating optimization pathologies caused by high gradients near the airfoil. Building on this, NNfoil-C incorporates flow condition parameters (Mach number and angle of attack) and 12 class shape transformation parameters controlling the airfoil shape into the network input. They combine with the spatial coordinates $x$, $y$ form 16 model inputs, corresponding to a 16-dimensional complete state-space. The model outputs include density $\rho$, velocity in the $x$-direction $u$, velocity in the $y$-direction $v$, and pressure $p$. The loss function of NNfoil-C includes equation loss and boundary condition loss, similar to PINNs, but due to the input design, NNfoil-C's loss controls all flows within the complete state-space. For more details, please refer to the original literature.

Compared to traditional numerical methods, NNfoil-C's advantage lies in its ability to obtain all flow fields within the complete state-space through a single, short-duration unsupervised training session. In contrast, traditional methods can only handle one flow condition or shape at a time, resulting in significant time costs when evaluating numerous flow conditions and shapes in engineering applications. NNfoil-C, however, completes this task in less than 20 hours.

\subsubsection{Inference and Perturbations}

\textbf{Inference Process.}
The inference process of NNfoil-C involves using the trained model to predict the pressure coefficient distribution on the airfoil surface for six different states and shapes designed in the original literature. The model training includes two stages: pre-training and fine-tuning. In the pre-training stage, the L-BFGS optimizer is used for 1250 steps, with a maximum inner iteration of 1000. The fine-tuning stage also uses the L-BFGS optimizer, executing 500 steps.

\textbf{Adversarial Attacks.}
To evaluate the vulnerability of the NNfoil-C model, we introduced perturbations to the spatial coordinates $X=[x,y]$ in the input of NNfoil-C for 200 training points on the airfoil surface. These perturbations included random noise and FGSM attacks, both with an amplitude of $X\epsilon(\epsilon=3\%)$, differing only in direction. The random noise had completely random directions, while the FGSM attacks followed the direction of the gradients of the loss with respect to $x$ and $y$. The remaining flow condition parameters and shape parameters in the input are kept unchanged. We input the perturbed data from six cases into the trained NNfoil-C model and evaluate the relative L2 error between the pressure coefficient distributions on the airfoil surface obtained by the model and the reference solutions obtained through the finite volume method, thereby completing the evaluation of NNfoil-C.

\subsection{Deep-Learning Quasi-Particle Model}

\subsubsection{Model Overview}

The deep-learning quasi-particle (DLQP) model is developed to extract useful information from the Lattice QCD equation of state of hot and dense nuclear matter under extreme conditions  \cite{Li:2022ozl}. Utilizing the concept of quasi-particles from mean-field theory, the strongly coupled quark gluon plasma is treated as weakly interacting quasi-parton gas. Here, 3 ResNet blocks are employed to represent masses of u/d quarks $m_{u/d}(T)$, s quarks $m_s(T)$ and gluons $m_g(T)$, all of which are functions of temperature $T$. These masses approximate the energy carried by these quasi-partons within the interaction range at various temperatures. Based on this approximation, the partition function $\ln Z$ as well as the QCD EoS are derived through straightforward statistical dynamics,

\begin{align}
\ln Z(T) &= \ln Z_\text{g}(T) + \ln Z_{\text{u/d}}(T) + \ln Z_\text{s}(T), 
\nonumber\\
P(T) &= T \left( \frac{\partial \ln Z(T)}{\partial V} \right)_{T}, 
\nonumber\\
\epsilon(T) &= \frac{T^{2}}{V} \left( \frac{\partial \ln Z(T)}{\partial T} \right)_{V},
\nonumber\\
s(T) &= \frac{\epsilon + P}{T},
\label{eq:lnz_total}
\end{align}
where $\ln Z_\text{g}(T)$, $\ln Z_{\text{u/d}}(T)$, and $\ln Z_\text{s}(T)$ are the partition functions for gluons, u/d quarks, and s quarks, respectively. These partition functions are computed using the momentum integration:

\begin{align}
\ln Z_\text{g}(T) &= - \frac{d_\text{g} V}{2 \pi^{2}} \int_{0}^{\infty} p^{2} dp \ln \left[ 1 - \exp \left( -\frac{1}{T} \sqrt{p^{2} + m_\text{g}^{2}(T)} \right) \right], 
\nonumber\\
\ln Z_{q_i}(T) &= + \frac{d_{q_i} V}{2 \pi^{2}} \int_{0}^{\infty} p^{2} dp \ln \left[ 1 + \exp \left( -\frac{1}{T} \sqrt{p^{2} + m_{q_i}^{2}(T)} \right) \right].
\label{eq:lnz_func}
\end{align}
where $d_\text{g} = 16$, $d_{q_\text{s}} = 12$, and $d_{q_{\text{u/d}}} = 24$ are the degrees of freedom for gluons and quarks, $V$ is the volume. 

Note that the derivatives in the formula are computed using auto differentiation and the one-dimensional integration is implemented numerically using Gaussian quadrature such that the integration will not hinder the back propagation process during training.
The training data is directly obtained from the HotQCD collaboration calculations \cite{HotQCD:2014kol}, where 50 points are selected that cover the transition temperature region.

After being trained, three mass functions for these quasi partons are obtained, which are useful for various studies. For instance, they are essential for calculating transport coefficients of quark-gluon plasma, such as shear viscosity, bulk viscosity and heat conductivity. This study provides a method to investigate the contributions of different parton flavors to these transport coefficients.
Furthermore, this study can be extended to finite baryon chemical potential in the future.

\subsubsection{Inference and Perturbations}

\textbf{Inference Process}

The inference process involves using the trained model to predict the pressure and energy density at different temperatures. In our example, we trained the model for 50,000 epochs to obtain the neural network parameters configuration. The trained neural networks can be conceptualized as three one-dimensional functions, where the input is temperature and the outputs are pressure and energy density.

\textbf{Adversarial Attacks}

To evaluate the robustness of the DLQP model under perturbations, we implemented adversarial attacks using the FGSM attack and random perturbations. Similar to previous tasks, the FGSM attack perturbs the input data by adding a small perturbation in the direction of the gradient of the loss with respect to the input. The random perturbation attack introduces a random noise tensor of the same magnitude as the FGSM perturbation. Since the input is a one-dimensional scalar, the random noise has a 50\% chance to coincide with that of the FGSM attack at each sampling. To better reflect the arbitrariness of random noise, we need to perform multiple samplings and compute the average error which is equivalent to the average error introduced by one positive and one negative noise.

During testing, we input the original temperature, the temperature after the FGSM attack, and the temperature with random noise into the neural network, respectively, to calculate the absolute error between the predicted pressure and the true pressure. Since the results of lattice calculations are range-limited, the selection of $\epsilon$ should not be too large, as it may lead to extrapolation over a large temperature interval where lattice data is lacking, making the comparison of results meaningless.

\subsection{Wireless transmission in communication}
 
We introduce the learning-based algorithm utilizing DNN for optimal beam prediction in user equipment (UE). The algorithm employs direct UE measurements, significantly enhancing the efficiency of beam selection. The process is divided into three main phases: Reference Signal Received Power (RSRP) measurement, training phase, and prediction phase. In the RSRP measurement phase, the UE measures the RSRP across a set of predefined beams (one can consider the RSRP as a numerical value received by the UE in order for communication). Measurements are conducted on two beam types: fine beams with narrow coverage and broad beams with wider coverage. The RSRP values comprise a tensor and are processed following a pilot normalization policy, which serves as input to the DNN model. The training phase involves training the model using an off-line reinforcement learning (RL) framework. The training data is collected from simulations aided by the parameters (i.e., the channel matrix) from the experiments, where the signal processing is conducted using an NI USRP-Rio software-defined radio platform (USRP) for baseband processing. The system implements critical baseband functions, including 5G NR frame structure and RSRP measurements. The USRP is controlled via a laptop and communicates bidirectionally using 28 GHz mm-wave frequency bands. These data are compatible with the New Radio (NR) standards or through a proof-of-concept (PoC) mm-wave beam management hardware prototype, aligned with the 5G standards. In the subsequent prediction phase, the trained DNN model predicts the optimal receive (Rx) beam based on the processed input, making the beam selection decision for the UE.

The input data for the machine learning model is structured as an $M \times (N + 1)$ matrix, where $N$ is the number of Rx beams and $M$ is the history length. Each matrix row contains $N + 1$ elements, representing RSRP values for the $N$ fine beams and one broad beam. Only one fine beam is measured per step, with other entries filled using a default RSRP value. Since we take 3 beams as input, here $N=3$. This matrix design aids the model in recognizing the measured Rx beam based on its matrix position. Data pre-processing is employed to enhance the training process. The RSRP data is normalized by scaling and bias adjustment, centering the values around zero. The normalized RSRP value, denoted as $E$, is computed as:
\begin{align}
E = \frac{1}{g}(\text{RSRP} + b),
\end{align}
where $b$ is the bias term and $g$ is the scaling factor. For decision-making, the beam management model utilizes a Q-learning algorithm, which ranks beam candidates based on Q-values, representing the expected long-term reward of each selection. The optimal beam is identified as the one with the highest Q-value.

The Q-learning model is defined with the following components: the state $s$ represents the UE's current RSRP measurements, the action $a$ denotes the selection of a different Rx beam, and the reward $r$ is determined by the observed improvement in RSRP. The objective is to maximize the cumulative discounted reward $Q$, given by:
\begin{align}
Q^*(s, a) = \max_{\pi} \mathbb{E}\left[r_t + \gamma r_{t+1} + \ldots \mid s_t = s, a_t = a, \pi\right],
\end{align}
where $\pi$ is the policy mapping state $s$ to action $a$, and $\gamma$ is the discount factor. The optimal action $a_t^*$ is selected as:
\begin{align}
a_t^* = \arg \max_{a_t \in A} Q(s_t, a_t; \theta),
\end{align}
with $\theta$ representing the parameters of the Q-function. Given the sequential nature and high dimensionality of RSRP data, a Recurrent Neural Network (RNN) is adopted for the Q-function. Specifically, a Long Short-Term Memory (LSTM) network is selected due to its ability to capture temporal dependencies and high-level features in sequential data. The LSTM network employs input, forget, and output gates to control information flow, effectively managing memory and transforming RSRP sequences into Q-value predictions. By leveraging the multi-layer LSTM model, the proposed algorithm efficiently predicts the optimal beam for the UE, demonstrating robust performance in dynamic radio environments. This approach enhances beam management and significantly improves overall network performance.

\subsection{Randomized neural network and DNN in solving Poisson equation}
\subsubsection{Task overview}
We solve the Poisson equation outlined below:
\begin{align*}
    -\Delta u&=f\quad\text{in }\Omega,\\
    u&=g\quad\text{on }\partial\Omega,
\end{align*}
where $\Omega=[-1,1]^2$ and the exact solution is $u(x,y)=\sin(4\pi x)\sin(4\pi y)$. 

\subsubsection{Randomized neural network for Solving the 2D Poisson Equation}
Randomized neural networks (RNN, \cite{SHANG2023107518,SUN2024115830}) differ from general neural networks in that most parameters are pre-set, and only the parameters of the last hidden layer require training. In RNN, the network can be seen as a linear combinations of basis functions, with the basis functions being the last hidden layer and the trained parameters acting as combination coefficients. In contrast to traditional neural network approaches, RNN simplifies the training process by solving a linear system using the least-squares method instead of the stochastic gradient descent method, resulting in reduced training time and improved numerical accuracy.

We use a shallow neural network consisting 400 neurons in the hidden layer, with the input being a two-dimensional vector and the output representing a scalar value. The fixed parameters are randomly selected from a uniform distribution $\mathcal{U}(-1,1)$. With 900 interior points and 2000 boundary points (distributed as $4*500$), and a boundary penalty set at 100, we proceed to solve a $2900\times400$ linear system through the least-squares method, with the solution determining the weights of the final layer.

To conduct adversarial attacks, we sample $101\times101$ uniform points for the purpose of attacking the RNN using the FGSM. In cases where the perturbed data lies outside the domain $\Omega$, we remove these data points.

\subsubsection{Deep Neural Network for Solving the 2D Poisson Equation}

The PINN is implemented as a deep neural network (DNN) with the following characteristics:
\begin{itemize}
    \item \textbf{Input}: A two-dimensional vector representing spatial coordinates \((x, y)\).
    \item \textbf{Output}: A scalar value representing the solution \(u(x, y)\) of the Poisson equation.
    \item \textbf{Architecture}: The network consists of 5 hidden layers, each with 100 neurons, followed by a single output neuron. The activation function for the hidden layers is the hyperbolic tangent (Tanh).
    \item \textbf{Normalization}: The input data is normalized to the range \([-1, 1]\) to improve training stability.
\end{itemize}

The Adam optimizer is used to minimize the loss function, with a learning rate of \(0.0008\). The training process continues until the loss falls below a specified threshold or a maximum number of steps is reached.

To evaluate the robustness of the PINN, we apply the FGSM to the test data with perturbation magnitude controlled by the parameter \(\epsilon\), which takes values \([0.01, 0.05, 0.1, 0.15, 0.2, 0.25, 0.3]\).

\backmatter

\bmhead{Acknowledgements}

We are grateful to the discusion of the vulnerability of DeepMD with Dr. Linfeng Zhang in AI for Science Institute and DP Technology.

\section*{Declarations}

\begin{itemize}
\item Competing interests: All authors declare no competing interests.
\item Code and data availability: The code and data will be provided by the corresponding author with reasonable request.
\item Author contribution: 

\textbf{Jun-Jie Zhang:} Designed the verification of the work, performed experiments of FourCastNet and DeepMD.

\textbf{Jian-Nan Chen:} Worked on the analysis of all models.

\textbf{Jiahao Song, Weiwei Zhang:} Performed the experiments of fluid dynamic problem.

\textbf{Xiu-Cheng Wang, Yiyan Zhang, Nan Cheng:} Performed the experiments of wireless communications.

\textbf{Fu-Peng Li, Long-Gang Pang:} Performed the experiments of Quantum Chromedynamics.

\textbf{Zehan Liu, Shiyao Wang:} Performed quantum chemistry calculations using VASP.

\textbf{Duo Zhang:} Worked on the vulnerability of DeepMD.

\textbf{Jianhui Xu, Chunxiang Shi:} Worked on the vulnerability of FourCastNet.

\textbf{Haoning Dang, Fei Wang:} Worked on the attack of Randomized neural network.

\textbf{Deyu Meng:} Supervised the work.

\end{itemize}

\bibliography{sn-bibliography}% common bib file

\end{document}